\documentclass[11pt,a4paper]{article}
\usepackage[hyperref]{acl2022}
\usepackage{times}
\usepackage{latexsym}
\usepackage{graphicx}
\usepackage[ruled,vlined]{algorithm2e}

\usepackage{float}

\makeatletter
\newcommand{\removelatexerror}{\let\@latex@error\@gobble}
\makeatother
\usepackage{microtype}

\aclfinalcopy 

\title{ACM - Attribute Conditioning for Abstractive Multi Document Summarization}

\author{Aiswarya Sankar \\
  UC Berkeley \\
  \texttt{aiswarya.s@berkeley.edu} \\\And
  Ankit Chadha \\
  Stanford  \\
  \texttt{ankitrc@stanford.edu} \\}

\date{}

\begin{document}
\maketitle
\begin{abstract}
Abstractive multi document summarization has evolved as a task through the basic sequence to sequence approaches to transformer and graph based techniques. Each of these approaches has primarily focused on the issues of multi document information synthesis and attention based approaches to extract salient information. A challenge that arises with multi document summarization which is not prevalent in single document summarization is the need to effectively summarize multiple documents that might have conflicting polarity, sentiment or subjective information about a given topic. In this paper we propose ACM, attribute conditioned multi document summarization,a model that incorporates attribute conditioning modules in order to decouple conflicting information by conditioning for a certain attribute in the output summary. This approach shows strong gains in ROUGE score over baseline multi document summarization approaches and shows gains in fluency, informativeness and reduction in repetitiveness as shown through a human annotation analysis study.
\end{abstract}

\section{Introduction}

Abstractive multi document summarization is the task of writing a single summary of the key points and content in multiple related documents. This task has evolved from research in single document abstractive and extractive summarization; however, it faces unique challenges due to input documents having duplicate and conflicting content as well a larger body of text. \cite{radev-etal-2000-centroid}. This task has evolved from early approaches using sequence to sequence (Seq2Seq) neural architectures to transformer based architectures with the introduction of large-scale datasets \cite{DBLP:conf/iclr/LiuSPGSKS18}, \cite{fabbri2019multi}. Beyond the introduction of approaches now commonly used for single document abstractive summarization, cross document attention and graphs that capture relations between text in various documents have further improved the state of the art for multi document summarization tasks. \cite{DBLP:conf/iclr/LiuSPGSKS18}, \cite{DBLP:conf/emnlp/LiLLZW14}. These graphs aim to better represent the inter dependencies between articles by representing text spans as nodes in the graph and capturing the relations between these sentences as edge weights. 

Despite the advances made with these approaches, a significant challenge remains in multi document summarization with respect to how to deal with contradictory information present in the multiple source documents. It is critical to both learn the relationships between different documents as well as to extract salient information that is consistent with the output viewpoint. This is a situation often faced with summarizing multiple news articles where different viewpoints on an issue can significantly change the semantic structure of the content present in each article making it challenging for the abstractive summarization model to learn the relationships between inconsistent or conflicting information. In this work we define conflicting opinions as a combined measure of the polarity and sentiment of text. By this definition, two pieces of text on the same topic that have a differing sentiment and/ or polarity are determined to have different viewpoints. This definition is used throughout the paper. 

\begin{table}[ht]
\centering
\begin{tabular}{l}
\hline \textbf{Source 1}\\ \hline
A fresh update on the U.S. employment situation \\for January  hits the wires at 8:30 a.m. New York \\time offering one of the most important snapshots \\on how the economy fared during the previous \\month. \textbf{Expectations are for 203,000 new jobs to}\\  \textbf{be created. The unemployment rate is expected} \\\textbf{to hold steady at 8.3\%}.\\
\hline \textbf{Source 2}\\ \hline
Employers \textbf{pulled back sharply on hiring last} \\\textbf{month, a reminder that the U.S. economy}\\ \textbf{may not be growing fast enough} \\to sustain robust job growth. \textbf{The unemployment}\\ \textbf{rate dipped}, but mostly because more Americans\\ stopped looking for work.\\
\hline
\end{tabular}
\caption{\label{DataComp1} Contradictory information in source articles poses a challenge when generating a multi document summary. }
\end{table}

This paper proposes ACM, attribute conditioned multi document summarization, a novel approach to multi document summarization that incorporates an attribute conditioning module with an abstractive multi document summarization model in order to condition for a particular attribute when generating the multi document summary. This approach addresses the challenge of dealing with conflicting information in the input documents by conditioning for a particular attribute in the input text, a problem not yet addressed in the literature. Table 1 presents an example of input documents with conflicting information. The attribute conditioning model is trained in order to decouple conflicting information as defined by different sentiment and polarity attributes, however this module is highly extensible and can address other attributes as well. In this paper, we choose polarity and sentiment as human analysis has shown they closely capture opinions in text. We train these classification models on every input prefix (n-gram) in the input dataset in order to train the model to be agnostic to input text length. We evaluate these approaches on baseline abstractive multi document summarization architectures in order to observe improvements in the output as evaluated through ROUGE metrics and through human annotations for fluency, informativeness and consistency. Our approach consists of individual composable elements, each of which we further evaluate independently through ablation studies. One limitation of this work, however, includes being able to identify attributes that best capture the nature of the conflicting information in the text as this would be different based on each dataset used. For example given movie reviews, an attribute such as polarity would likely not provide a significant improvement when generating the multi document summary. Thus the attribute needs to be carefully selected and fine tuned based on the source of documents in order to identify different classes of information. 

\begin{itemize}
    \item{Graph conditional weighting - We learn a graphical representation of the input documents that weights graph edges by incorporating both the conditional attribute score for each input as well as the cosine similarity between the paragraphs.}
    \item{Conditional training - We conditionally fine tune the abstractive multi document summarization module by combining the logits for each input prefix passed to the decoder module with the conditioning score for that prefix from the attribute conditioning module.}
    \item{Attribute future discriminators - When evaluating the model, we modify beam search to rank each beam according to the product of the attribute conditioning model and the conditional likelihood score.}
\end{itemize}   

The contributions of our work are as follows:
\begin{itemize} 
\item{We propose a novel attribute conditioning layers that can be used with graph-based and transformer based layers for MDS aimed to improve view point consistency for the task of multi-document summarization.}
\item{We provide an evaluation on ROUGE and human annotation analysis of the output summaries evaluating for informativeness, consistency and fluency on the MultiNews dataset.}
\item{We perform an ablation study to analyze contribution of each composable component in our proposed architecture.}
\end{itemize}

\section{Related Work}

This paper builds on techniques developed through the following NLP tasks - abstractive summarization, multi document summarization, and conditional language modeling - to efficiently address the issue of decoupling conflicting attributes in multi document summarization. We define the task of incorporating conflicting attributes e.g. sentiment and polarity as defined by a classification model in abstractive summaries as an application of conditional language modeling techniques. Conditional language modeling is a key factor in this problem as we aim to decouple the conflicting information in the summaries by conditionally selecting more coherent and consistent summaries.

\subsection{Conditional Language Modeling}

Conditional language modeling has been approached both through applying global constraints on text generation, by applying control only at inference time and by directly optimizing through policy gradient methods. \cite{ranzato2016sequence}, \cite{holtzman2018learning}, \cite{Dathathri2020Plug}. Alternative approaches include relying on predefined sets of control tokens or control codes as a form of a copy mechanism \cite{luong2015addressing}. These approaches have been successful at steering language models towards specific features or as in \cite{Dathathri2020Plug} with conditioning for the outputs to include words conditioned for through a simple bag of words attribute model. We build on these approaches to design the attribute conditioning module.

\subsection{Abstractive Summarization}
Abstractive summarization (AS) is a task in which a language model is trained to generate text that matches a pre-generated summary for a given article.  AS has gone through several phases of which the pioneering work was carried out by \cite{DBLP:journals/corr/SutskeverVL14} where an Seq2Seq RNN Model was implemented to generate text. The Seq2Seq RNN Model inherently had multiple challenges such as altering factual details and redundancy. \cite{DBLP:journals/corr/SeeLM17} circumvented the issues by creating a pointer-generator model which keeps track of words and sequence in the original text and using them in the result hence ensuring the meaning of summary is in-line with the original text. This paper also included a coverage mechanism to keep of track of which parts of the original text have been summarized thus penalizing repetition. BART, the bi-directional auto-regressive transformer \cite{DBLP:journals/corr/abs-1910-13461}, built on this work by improving on the task of abstractive summarization by introducing arbitrary noise in the input text and training the model to reconstruct the original text. 

\subsection{Multi Document Summarization}

Multi document summarization has evolved through four primary approaches since the task was first introduced.  The first set of approaches focused on graph ranking based extractive methods through TextRank {\cite{mihalcea-tarau-2004-textrank}}, LexRank {\cite{DBLP:journals/jair/ErkanR04}} and others. These approaches came before syntax and structure based compression methods which aimed to tackle issues of information redundancy and paraphrasing between multiple documents. Compression-based methods as shown in {\cite{DBLP:conf/emnlp/LiLLZW14}} and paraphrasing based were improved upon with the advent of neural seq2seq based abstractive methods in 2017. This allowed multi document summarization to further improve upon the work done with single document abstractive summarization through approaches such as pointer generator- maximal marignal relevance {\cite{lebanoff2018adapting}}, T-DMCA {\cite{DBLP:conf/iclr/LiuSPGSKS18}} the paper that also introduced the foundational WikiSum dataset and HierMMR {\cite{fabbri2019multi}} that introduced MultiNews. These approaches aimed to tackle information compression through maximal marginal relevance scores across documents and through attention based mechanisms. 

Improvements upon those baseline models include further leveraging graph based approaches to pre-synthesize dependencies between the articles prior to multi document summarization as tackled in {\cite{li-etal-2020-leveraging-graph}}. Further work needs to be done to further exploit these graphical representations as {\cite{li-etal-2020-leveraging-graph}} essentially works to establish baselines with tf-idf, cosine similarity and a graphical representation first described in {\cite{christensen-etal-2013-towards}}. These papers primarily aim to address de-duplicating information and learning relationships between the different topics shared across documents however none of these architectures are built to deal with conflicting information.

\section{Our Approach}

We present a novel technique ACM, attribute conditioned multi document summarization, which is designed to address the problem of resolving conflicting information in multi document summarization through the use of an attribute conditioning module. Conflicting information is determined by the attribute conditioning model trained on both a polarization and a sentiment analysis dataset since both factor into determining information consistency. XLNet {\cite{yang2019xlnet}} is used as the model architecture for the attribute conditioning module and is trained using sentence prefixes in order to capture both word level and phrase or sentence level features. XLNet has shown state of the art results in sentiment analysis tasks. The outputs of this classifier are used in each approach in order to fine tune the model to consistently condition for a particular attribute.  

\begin{figure*}[h]
\includegraphics[scale=.5]{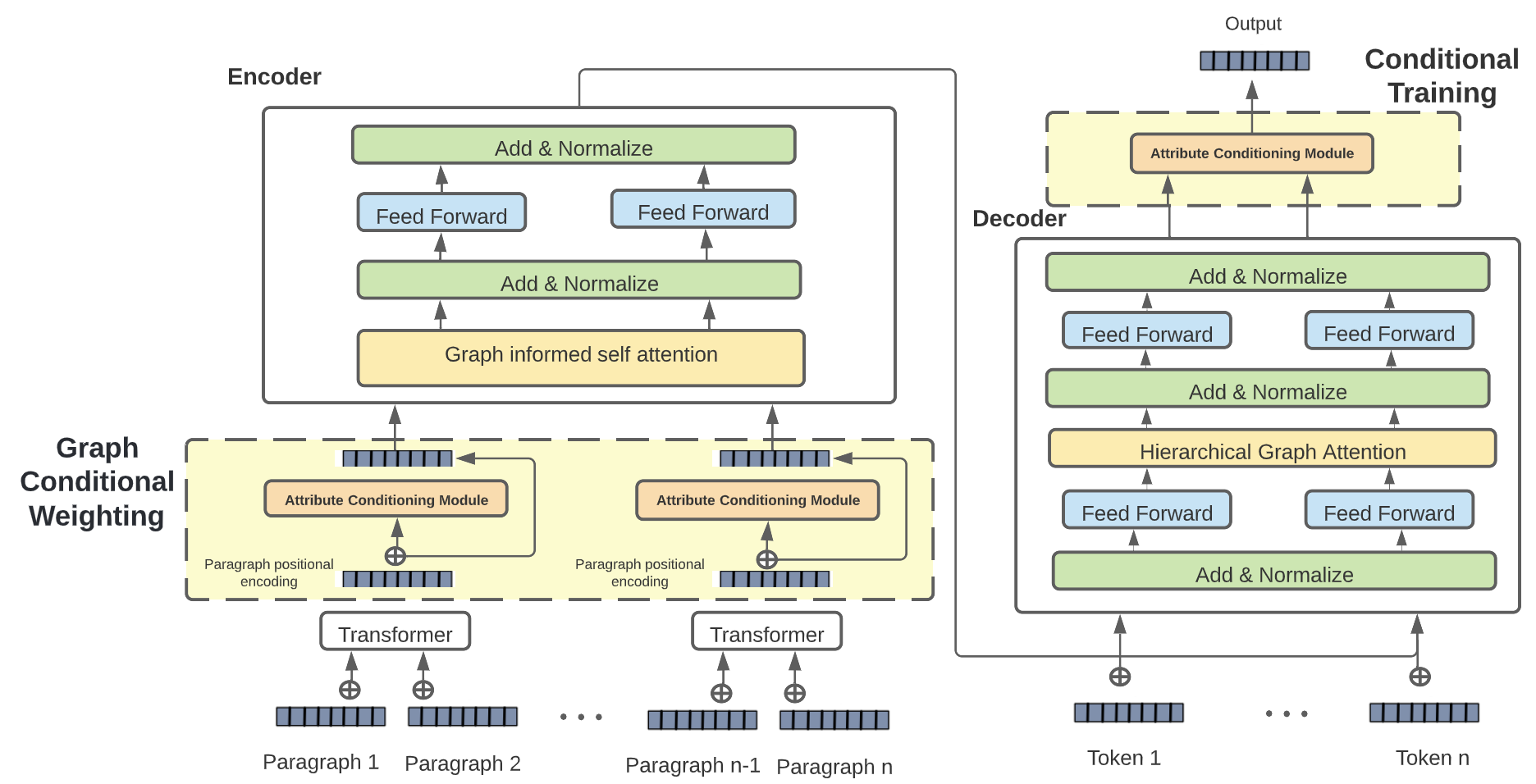}
\caption{Attribute conditioned multi document summarization (ACM) model diagram. At each stage of the summarization process, ACM uses the attribute conditioning module to preserve viewpoint consistency in the output summary. As attribute future discriminators is a component of model evaluation, it is included separately in figure 3. }\label{fig:Future_discriminators}
\end{figure*}

\begin{figure}
\removelatexerror
\begin{algorithm}[H]
\SetAlgoLined
\DontPrintSemicolon
\KwResult{Conditional MDS with Future Discriminators}
beams = []\;
beamWidth = 200\;
scores = [];\;
 \For{t = 1...T}{\;
    topBeams = argmax(beams, beamWidth);\;
    \For{beam $\in$ topBeams } {\;
        topNextTokenLogits = argmax(nextTokenLogits, 200)\;
        topNextTokens = vocabToIndex(topNextTokenLogits)\;
        combinedScores = []\;
        
        \For {token, tokenLogits $\in$ (topNextTokens, topNextTokenLogits) } {\;
            beam' = beam + token\;
            \textbf{conditionalLogits = conditionalModel(beam')}\;
            \textbf{combinedScores += conditionalLogits*topNextTokenLogits}\;
            beams = beams $\bigcup$ beam'\;
        }
    }
 }
\end{algorithm}

\begin{center}
\caption{Future discriminators for MDS beam search incorporates the attribute conditioning model at evaluation time in order to weight each beam based on the combined attribute score and maximum log likelihood.}
\end{center}
\label{fig:Conditional_training}
\end{figure}

In order to preserve viewpoint consistency, the attribute conditioning module is used to guide each stage of the summarization process. To prevent over fitting to the conditioning model, each stage is conditioned with a weighting term set as a hyper parameter when training ACM.  We also include each approach as an ablation study in order to determine how effective each stage of conditioning is on the final output summaries. In these ablation studies, we aim to investigate how well the attribute conditioning model can show improvements over three different baseline abstractive multi document summarization models - BART {\cite{DBLP:journals/corr/abs-1910-13461}}, BART with longformer self attention {\cite{DBLP:journals/corr/abs-2004-05150}} and GraphSum {\cite{li-etal-2020-leveraging-graph}}.

\subsection{Graph conditional weighting layer}

We preprocess the inputs to the model through a graph conditional weighting layer. As shown in Figure 1, for each of the input paragraph vectors both the positional encoding layer and the attribute conditioning layer outputs are computed. The graph representation is then constructed by creating a matrix of values where each row and column represents a paragraph in one of the input documents. The baseline model for GraphSum {\textcolor{blue}{\cite{li-etal-2020-leveraging-graph}}} computed a similarity score between these paragraphs by using tf-idf. Our graph conditional weighting layer aims to improve this approach by adding the weighted sum of linearly transformed paragraph vectors with the product of the attribute conditioning module for each paragraph as represented by $\beta_i$ and $\beta_j$ incorporating similarity between the attribute scores for each paragraph. 

$$\alpha_{ij} = softmax(e_{ij} + R_{ij}) $$
$$e_{ij} = \frac{(x_i^{l-1}W_{Q})(x_j^{l-1}W_K)^T }{\sqrt{d_{head}}} + \beta_i * \beta_j  $$
$$u_i = \sum_{j=1}^{L}\alpha_{ij}(x_j^{l-1}W_V) $$
$$R_{ij} = -\frac{(1 - G[i][j])^2}{2\sigma^2} $$

Let $x_i^{l-1}$ represent the output of the graph encoding layer for paragraph $x_i$. Let $R_{ij}$ represent the pairwise relation bias of the weights of graph representation matrix $G$. Then graph informed self attention with conditional weighting can be defined as taking the product of the attribute conditioning scores between each pair of paragraphs with the weighted sum of the linearly transformed paragraph vectors.  Thus the graph will learn stronger weights between paragraphs that are of the same polarity or sentiment when generating the output summary. 

\subsection{Attribute future discriminators}
Attribute future discriminators approaches the problem with Bayesian factorization - effectively decoupling the pre-trained summarization model from the fine tuning classification model. This approach is described as using a future discriminator as it takes the sequence of text generated so far, appends to it the most likely next token predicted by the summarization model and then computes the likelihood that this sequence satisfies the desired classification attribute. In doing so it increases the likelihood of both generating text that satisfies the summarization model with a high likelihood as well as staying consistent with the desired text modeling attributes. 

We can model the standard summarization task as 
$${P(X) = \Pi_i^n P(x_i|x_{1:i-1})}$$
Let $X$ refer to a set of input documents $d_1,..d_n$ pertaining to a specific MDS summary $s_i$. Let each document $d_i = {x_{i1},..x_{in}}$ where $x_{ij}$ refers to token $j$ in document $i$. Prior to the MDS task, all input documents are appended, thus absolving the need to index each token with its corresponding document. Lastly, let $a$ refer to the conditioning attribute. 
When conditioning on a certain attribute, we model the task as $$P(X|a) = \Pi_i^n P(x_{1:i-1},a)$$
We rely on the following factorization to decouple the summarization module from the attribute conditioning model:
$$P(X|a) \propto P(a|x_{1:i}) P(x_i|x_{1:i-1})$$ 
This approach allows us to train the conditional model and the summarization model separately and in a composable manner to achieve the desired output conditioning as shown in figure 2.

As both the summarization model and the classification models are pretrained, during evaluation each of the top k words selected by the decoder logits in the summarization model is added to the best sequence so far and then passed into the attribute classification model. The output probabilities of both models are then combined in selecting the next best word generated by the combined model. This allows for a high degree of composability as each of the attribute models can be layered on top of each other and added to the final logits with different weights.

\subsection{Conditional training}

Conditional training combines the probability distribution over the next words generated by the decoder with the polarity scores as determined by the attribute classification model. In contrast to other models such as {\textcolor{blue}{\cite{Dathathri2020Plug}}}, this allows the base model to be conditioned to output based on the desired attribute similar to the approaches taken to train class conditional language models. 

For each model architecture, the decoder logits are multiplied with the attribute conditioning model's logits during training. This trains the abstractive multi document summarization model to output text conditioned on that particular attribute. Thus for each of the input documents, information that aligns with the selected attribute is proportionally more likely to be generated. The same classifier models and baseline MDS models were used to evaluate this approach as in the previous method. We show that this approach can be applied to any baseline MDS architecture in the follow up ablations.

\section{Experiments}
\subsection{Datasets}
In order to determine conflicting information in text, we first train attribute classification models on two different datasets AllTheNews {\cite{Thompson2020allthenews}} and the MPQA Opinion Corpus dataset {\cite{deng-wiebe-2015-mpqa}} in order to determine the sentiment and polarity scores of input text. These attributes are strongly correlated through human analysis with differing viewpoints on text. In order to create a dataset most similar to the input sequences that would be passed through ACM, the input text in the dataset was augmented to include all prefix length subsequences with the same label. This corresponds to the prefix length subsequences that will be evaluated for the above approaches. 
   
AllTheNews dataset consists of 2.7 million articles from 26 different publications ranging from January 2016 to April 2020 in English{\cite{Thompson2020allthenews}}. This dataset was augmented with polarity labels according to the news source label. The MPQA Opinion Corpus was chosen over other sentiment analysis datasets as it consists of articles pulled from news articles on a broad range of news sources and is consistent with the approach used in {\cite{ramsauer2021exploring}}.

The MPQA Opinion Corpus is copyrighted by MITRE Corporation and consists of 700 English articles from a range of English language sources. The dataset provides labeled entity spans with the corresponding positive or negative annotations. {\cite{deng-wiebe-2015-mpqa}}

MultiNews {\cite{fabbri2019multi}} consists of a varied set of news articles spanning over 1500 sites and their corresponding human written summaries. The average length of the source documents is 2100 tokens which allows the Bart+Longformer model trained with input length 4096 tokens to consume multiple concatenated articles at a time. For all models, each of the input documents are concatenated together and fed into into the model in batches of the maximum token size. XLNet was used as the primary architecture for each of the classification modules. Each of the attribute classification modules were trained with a 80-10-10 train, val, test split using the entire corpus of data. Classification results were averaged over three runs. 

\section{Results}

\begin{table*}[h]
\centering
\begin{tabular}{l}
\hline \textbf{Input Text} \\ \hline
\textbf{Article 1} \\
The court’s return to the subject after an appeals court sustained the hybrid plan has struck many supporters \\of {{\color{red}affirmative action as an ominous sign.}} Supporters worry Obama's efforts to {{\color{red}expand his executive actions}} \\ {{\color{red}on immigration will remain blocked by a lower court order.}}\\
\textbf{Article 2} \\
The cases on {{\color{blue}unions and affirmative action}}, for instance, were almost certainly added to the docket \\ \color{blue}{by the more conservative justices in the confidence that they would be able to move the law to the right}\\ allowing religious institutions to opt out of providing Obamacare.\\
\textbf{Article 3} \\
An appeals court has upheld most such limits in Texas, but in June the Supreme Court voted, \color{orange}{5-4, to }\\\color{orange}{block key parts of the law until the justices decide whether to weigh in.}\\

\hline \textbf{GraphSum Baseline} \\ \hline
The Supreme Court's new term kicks off Monday, but \color{red}{it's unclear whether the fight will get to the justices in} \\{{\color{red}time for a decision this term.}} The justices are already facing a batch of petitions involving the rights \\of {{\color{red} religious institutions to opt out of providing contraception under Obamacare,}} reports POLITICO. \\
\hline  \textbf{ACM} \\ \hline
The Supreme Court is facing a \color{blue}{ docket of high-profile political cases that will test whether recent} \\{{\color{blue}liberal victories were more fluke or firm conviction,}} the New York Times reports. The court— \\{{\color{orange}which is divided 5-4 for conservatives}}, but saw Justice Roberts vote liberal on Obamacare and same-sex\\ marriage —will look at cases including {{\color{red}unions, affirmative action, and possibly abortion.}}\\ 
\hline
\end{tabular}
\caption{\label{DataComp} This table provides an example provided for human annotation showing the comparison between GraphSum and ACM generated summaries given a set of input articles. Text highlighted in orange, red and blue show pieces of information extracted from the input articles that are present in the output summaries. ACM is able to effectively summarize opinionated content in the input summaries.}
\end{table*}

\subsection{ACM Training Configurations}
ACM was trained using 8 transformer encoder heads and 6 graph decoding layers. Beam size was set to 5 with length penalty factor 0.6 trained with gradient accumulation every 4 steps. Additionally we ran a hyper parameter search to determine the ideal weighting terms for the attribute conditioning module at each phase of summarization. For the ablation studies, BART and BART+Longformer self attention models were trained using the same set of hyperparameters used to train the baseline models. All models were trained using 2 NVIDIA K-90 GPUs over the span of ~1000 hours. BART, BART+Longformer, XLNet, and GraphSum each have 406M, 555M, 110M, and 121M parameters. Due to the composable attribute classifier modules in ACM, the number of parameters in ACM ranges from 130M to 240M including one attribute classifier. BART was trained with 8 transformer encoder heads and 6 decoder layers with gradient accumulation every 4 steps. The BART model was trained with max token size of 512 and the Bart+Longformer self attention model was trained with max token size of 4096. GraphSum {\cite{li-etal-2020-leveraging-graph}} was trained using the same configurations present in the original paper in order to reproduce results with 8 transformer encoder heads and 6 graph decoding layers. To maintain consistency with ACM, beam size was set to 5 with length penalty factor 0.6 trained with gradient accumulation every 4 steps. We performed hyper parameter search to determine the optimal weights for each of the modules. Let $\alpha_1$, $\alpha_2$ and $\alpha_3$ represent the weighting terms for each module, attribute conditioning with future discriminators, graph conditional weighting and conditional training, respectively. We found that $\alpha_1 = 0.22$, $\alpha_2 = 0.4$ and $\alpha_3 = 0.01$ showed the strongest ROUGE results. 
We evaluate our model on both ROUGE scores as well as perform human annotations to evaluate output summaries for fluency, informativeness, and consistency {\cite{lin-och-2004-automatic}}. ROUGE score is a standard metric for measuring summarization quality - ROUGE-1 measures overlap of unigrams, ROUGE-2, the overlap of bigrams and ROUGE-L, the longest common sub-sequence between the generated summaries and the gold standard references. ROUGE-L is often used as a measure of assessing fluency. {\cite{li-etal-2020-leveraging-graph}}. In addition we perform a series of ablation studies for each technique presented in order to assess the contribution of each to the final result.

\subsection{Evaluation Results}

We present an example summary highlighting the difference between the summaries generated with ACM versus the highest performing GraphSum baseline model in figure 2. These examples were provided for human annotation as is including the input articles as well as the GraphSum baseline and the ACM model summary. We primarily compare ACM against other strong baseline model architectures including BART \cite{DBLP:journals/corr/abs-1910-13461}, BART+Longformer attention \cite{DBLP:journals/corr/abs-2004-05150} and GraphSum \cite{li-etal-2020-leveraging-graph}. \\

\begin{table}
\centering
\begin{tabular}{|l|r|r|r|}
\hline \textbf{Model} & \textbf{R-1} & \textbf{R-2} & \textbf{R-L} \\ \hline

BART & 48.54* & 18.56* & 20.84*  \\
BART+Longformer & 49.03* & 19.04* & 24.0* \\
GraphSum & 42.99* & 27.83* & 36.97*  \\
ACM w/ Sentiment & 50.05 & 27.98 & 38.10 \\
ACM w/ Polarity & \textbf{50.12} & \textbf{28.12} & \textbf{38.19} \\
\hline
\end{tabular}
\caption{\label{DataComp2} Evaluation results for MDS graph with conditional weighting evaluated on the MultiNews dataset with ROUGE scores. Stared numbers are reproduced from the original papers.}
\end{table}

We evaluate each of these methods on the MultiNews dataset. MultiNews and WikiSum are the most commonly used datasets for multi document summarization. Since this paper is primarily concerned with addressing conflicting opinions, an entirely objective dataset such as WikiSum would not be ideal. Currently, there are no other large scale multi document summarization datasets from other domains. Table 3 shows the ROUGUE scores for each of the models. The first block of results represents the baseline models' ROUGE scores followed by the results from ACM. Overall performance shows that the sentiment attribute module and the polarity attribute module perform on par with polarity performing slightly better. This trend holds across the other approaches as well. ACM outperformed each of the baselines as well confirming our hypothesis that learning the relationships between the input data improves attribute consistency in the final output summary.

\subsection{Human Evaluation}

In addition to the automatic evaluation reported above, we also perform a human evaluation study with 5 evaluators on Amazon Mechanical Turk. We randomly select 182 input test summaries from the MultiNews dataset and the corresponding output summaries generated by ACM. In order to assess inter-annotator agreement, a portion of the articles were held out for annotation from each of the annotators. This analysis showed a 78\% agreement in scores amongst the annotators. Additionally, annotators were required to be fluent in the language.  In order to assess the quality of the model irrespective of the classifier model chosen, we randomly selected output summaries between the two classifiers. Since it is the previous state of the art, we use GraphSum as the baseline model.  
\begin{table}[h]
\centering
\begin{tabular}{|l|r|r|}
\hline \textbf{Model} & GraphSum & ACM  \\ \hline
Fluency & 3.48 & \textbf{3.91} \\
Repetitiveness & 2.3 & \textbf{1.74} \\
Informativeness & 3.24 & \textbf{3.86} \\
\hline
\end{tabular}
\caption{\label{DataComp3} Evaluation results from a human annotation study over 182 randomly selected output MDS summaries show strong improvements over the baseline model with respect to fluency, informativeness and repetitiveness on a scale of 1 to 5.}
\end{table}

\begin{table*}
\centering
\begin{tabular}{|l|r|r|r|}
\hline \textbf{Baseline Model} & \textbf{Rouge-1} & \textbf{Rouge-2} & \textbf{Rouge-L}  \\ \hline
BART & 48.54 & 18.56 & 20.84  \\
BART+Longformer & 49.03 & 19.04 & 24.0 \\
GraphSum & 42.99* & 27.83* & 36.97* \\
\hline
\hline \textbf{Our Approach} &  &  &  \\ \hline
BART w/ graph conditional weighting & 48.61 & 19.06 & 21.01 \\
BART w/ conditional training & 49.10 & 19.52 & 20.94 \\
BART w/ attribute future discriminators & 49.14 & 19.63 & 20.94 \\
BART+Longformer w/ graph conditional weighting & 49.32 & 19.34 & 24.42 \\
BART+Longformer w/ conditional training & 49.72 & 19.55 & 24.52 \\ 
BART+Longformer w/ attribute future discriminators & \textbf{49.83} & 19.85 & 24.32 \\ 
GraphSum w/ graph conditional weighting & 43.52 & 28.12 & 38.19 \\
GraphSum w/ conditional training & 44.71 & \textbf{28.22} & \textbf{38.61} \\
GraphSum w/ attribute future discriminators &  48.61 & 27.95 & 37.41 \\
\hline
\end{tabular}
\caption{\label{DataComp4} Ablation study comparing each of the approaches against baseline BART, BART+Longformer and GraphSum models. }
\end{table*}

Annotators assess the overall quality of the summaries based on three different criteria: (1) \emph{informativeness}, (2) \emph{fluency} and (3) \emph{repetitiveness}. Informativeness is defined as the number of unique facts / pieces of information present in the summary. Fluency is defined as the readability of the text accounting for good grammar, noun phrases and logical flow of information. Repetitive content comes from repeated words, phrases, or ideas throughout the output summary.  Each of these attributes were assessed on a scale of 1 to 5. According to this scale, a high score is preferable for fluency and informativeness and a lower score is preferable for repetitiveness.  A summary of the results from the study can be found in table 2. 

\subsection{Model Analysis and Ablations}

In order to determine the contribution of each method used within the ACM model, we performed additional ablations and model analysis. The key ablation studies included evaluating each approach on one of the baseline models, BART, BART + Longformer and GraphSum. We analyzed the results here both in terms of achieving higher ROUGE scores as well as maintaining information consistency. 

Table 5 presents the ROUGE scores for each of the approaches. We note that there are improvements from each approach individually with respect to the ROUGE score with attribute future discriminators performing marginally better than the other approaches. BART+Longformer achieves overall better performance on ROUGE scores as compared to BART primarily due to the longer input sequence lengths passed in. In addition to evaluating on ROUGE, an analysis is done on how well each approach was able to preserve the overall attribute conditioning.

\begin{table}[h]
\centering
\begin{tabular}{|l|r|r|}
\hline \textbf{Approach} & \textbf{Mean} & \textbf{Std Dev} \\ \hline
Graph conditional weighting & 0.76 & 0.288 \\
Conditional training & 0.82 & 0.274  \\
Attribute future discriminator & \textbf{0.891} & \textbf{0.103}  \\
\hline
\end{tabular}
\caption{\label{DataComp5}  Summaries generated with MDS with attribute future discriminators show the highest mean polarity score and narrowest standard deviation.}
\end{table}

Additionally we conducted a qualitative analysis of how well each approach was able to condition for polarity and sentiment in the output summaries by evaluating the summaries through the trained attribute conditioning module as shown in Table 6. This shows strong out of domain analysis results as the attribute conditioning module for polarity was trained on a different dataset, AllTheNews, and evaluated on the MultiNews dataset for MDS. This ablation study shows that MDS with future discriminators is the strongest attribute conditioning model. Since the sentiment and polarity of articles as determined by the XlNet classifiers are used to compute the opinion of an article, we used these models to analyze the MultiNews dataset. One possible risk that arises through these approaches is potentially increasing hallucinations given over conditioning using the attribute conditioning module. In this work, this risk was mitigated by fine tuning the weight of the attribute conditioning module to provide a balance between maximizing ROUGE while conditioning for a particular attribute.

\section{Conclusion}

In this work, we present a novel approach ACM, attribute conditioned multi document summarization, that sets the new state of the art for multi document summarization. It tackles the challenge of addressing conflicting information in multi document summarization by conditioning for a desired attribute and preserving consistency in the final output summary.  Through the attribute future discriminators, we are able to compose different conditional attribute models with a pretrained MDS model during evaluation. To our knowledge, this is the first approach taken to effectively decouple conflicting information by conditioning for a certain attribute in the output summary. This approach sets the current state of the art in ROUGE score over baseline multi document summarization approaches and shows gains in fluency and informativeness as well as a reduction in repetitiveness as shown through a human annotation analysis study. 

\bibliography{acl2022}
\bibliographystyle{acl_natbib}

\appendix

\end{document}